\pdfoutput=1

\documentclass[11pt]{article}

\usepackage[final]{acl}
\usepackage{times}
\usepackage{latexsym}
\usepackage[T1]{fontenc}

\usepackage{enumitem}
\usepackage{url}
\usepackage{times}
\usepackage{latexsym}
\usepackage{amsmath}
\usepackage{titlesec}
\usepackage{amssymb}
\usepackage{graphicx}
\usepackage{multirow}
\usepackage{comment}
\usepackage{array}
\usepackage{url}
\usepackage{adjustbox}
\usepackage{algorithm}
\usepackage[noend]{algpseudocode}
\usepackage[T1,T2A]{fontenc}     
\usepackage{tipa}
\usepackage{accents}

\usepackage[T1]{fontenc}

\usepackage[utf8]{inputenc}

\usepackage{microtype}

\usepackage{inconsolata}

\usepackage{graphicx}

\usepackage{booktabs}       
\usepackage{amsfonts} 
\usepackage{color, colortbl}
\usepackage{subcaption}
\usepackage{xspace}

\definecolor{Gray}{gray}{0.85}
\definecolor{LightCyan}{rgb}{0.88,1,1}
\newcolumntype{a}{>{\columncolor{LightCyan}}r}

\newcommand*{\gpt}{GPT-4.1\xspace}
\newcommand*{\ofour}{o4-mini\xspace}
\newcommand*{\geminiTwo}{Gemini 2.0 Flash\xspace}
\newcommand*{\geminiTwoFive}{Genini 2.5 Flash \xspace}

\newcommand*{\afroxlmrSixty}{AfroXLMR-61L\xspace}
\newcommand*{\afroxlmr}{AfroXLMR\xspace}

\newcommand*{\ibom}{\textsc{Ibom}\xspace}

\newcommand*{\yoruba}{Yor\`ub\'a\xspace}

\title{Ibom NLP: A Step Toward Inclusive Natural Language Processing for Nigeria's Minority Languages}

\author{
Oluwadara Kalejaiye\textsuperscript{1}, 
Luel Hagos Beyene\textsuperscript{2,3}, 
David Ifeoluwa Adelani\textsuperscript{4,5,6}, 
\\
\textbf{Mmekut-Mfon Gabriel Edet\textsuperscript{7}, 
 Aniefon Daniel Akpan\textsuperscript{8}, 
 Eno-Abasi Urua\textsuperscript{9}, 
 Anietie Andy\textsuperscript{1}}
\\ \\
 \textsuperscript{1}Electrical Engineering and Computer Science, Howard University, \\
 \textsuperscript{2}AIMS Research and Innovation Centre,
 \textsuperscript{3}NM-AIST, \textsuperscript{4}Mila - Quebec AI Institute, \\
 \textsuperscript{5}McGill University,
 \textsuperscript{6}Canada CIFAR AI Chair,
 \textsuperscript{7}Korapay,
\\
 \textsuperscript{8}National Institute for Nigerian Languages,
 \textsuperscript{9}University of Uyo
\\
 \small{
\textbf{Correspondence:} \href{mailto:email@domain}{anietie.andy@howard.edu}
 }
}

\begin{document}
\maketitle
\begin{abstract}
Nigeria is the most populous country in Africa with a population of more than 200 million people. More than 500 languages are spoken in Nigeria and it is one of the most linguistically diverse countries in the world. Despite this, natural language processing (NLP) research has mostly focused on the following four languages: Hausa, Igbo, Nigerian-Pidgin, and \yoruba (i.e $<$ 1\% of the languages spoken in Nigeria). This is in part due to the unavailability of textual data in these languages to train and apply NLP algorithms. 
In this work, we introduce \ibom---a dataset for machine translation and topic classification in four Coastal Nigerian languages from the Akwa Ibom State region: Anaang, Efik, Ibibio, and Oro. These languages are not represented in Google Translate or in major benchmarks such as Flores-200 or SIB-200.
We focus on extending Flores-200 benchmark to these languages, and further align the translated texts with topic labels based on SIB-200 classification dataset. Our evaluation shows that current LLMs perform poorly on machine translation for these languages in both zero-and-few shot settings. However, we find the few-shot samples to steadily improve topic classification with more shots. 






\end{abstract}

\section{Introduction}

Significant progress has been made towards developing and applying Natural Language Processing (NLP) and Machine Learning (ML) algorithms 
for translating textual data for low resource African languages \cite{kuwantomitigating, nwafor2022survey,adelani2022masakhaner, adelani2022few}. However, so far, these NLP and ML algorithms have been applied to only a few low-resource African languages. This is in part due to the unavailability of textual data in some of these languages~\citep{adelani-etal-2021-masakhaner}. Some African languages are not written; instead, they are orally passed down from one generation to the next, and they are not part of the educational system in their respective countries. The languages that receive attention are typically the most widely spoken, official or national languages~\citep{adelani2025natural}, which largely coincide with the top 50 African languages included in massively multilingual datasets~\cite{nllb2022,conneau2023fleurs,adelani-etal-2024-sib}.

Colonialism is in part responsible for the exclusion of these languages from the educational system.~\footnote{\url{hhttps://www.goethe.de/prj/zei/en/art/22902448.html}}  In the colonial times, only a few languages were 
encouraged in the educational system in the colonized countries; thereby ensuring that these languages were considered prestigious in comparison to other languages. Even when the colonized countries became independent, this practice was maintained in the respective countries. This has led to the endangerment and near extinction of some of these languages.


Nigeria is the most populous African country, with a population of more than 200 million people \footnote{\url{https://datacommons.org/place/country/NGA}}. There are more than 500 languages spoken in Nigeria \footnote{\url{ www.ethnologue.com}},
 making Nigeria one of the most linguistically diverse countries in the world \footnote{\url{https://www.weforum.org/stories/2023/04/worlds-most-multilingual-countries/}}. Despite the large number of languages spoken in Nigeria, very few of these languages have received sufficient attention as it relates to documentation and description. Nigerian languages are classified as either (a) "major" / "majority" or (b)  "minor" / "minority". The "majority" languages are Hausa, Yoruba, and Igbo. These majority languages have been taught in schools for decades and have significantly been documented; however, in comparison, the "minority" languages have received scant attention, and little has been documented in these languages.                             


\begin{figure*}[h]
\includegraphics[scale=0.58]{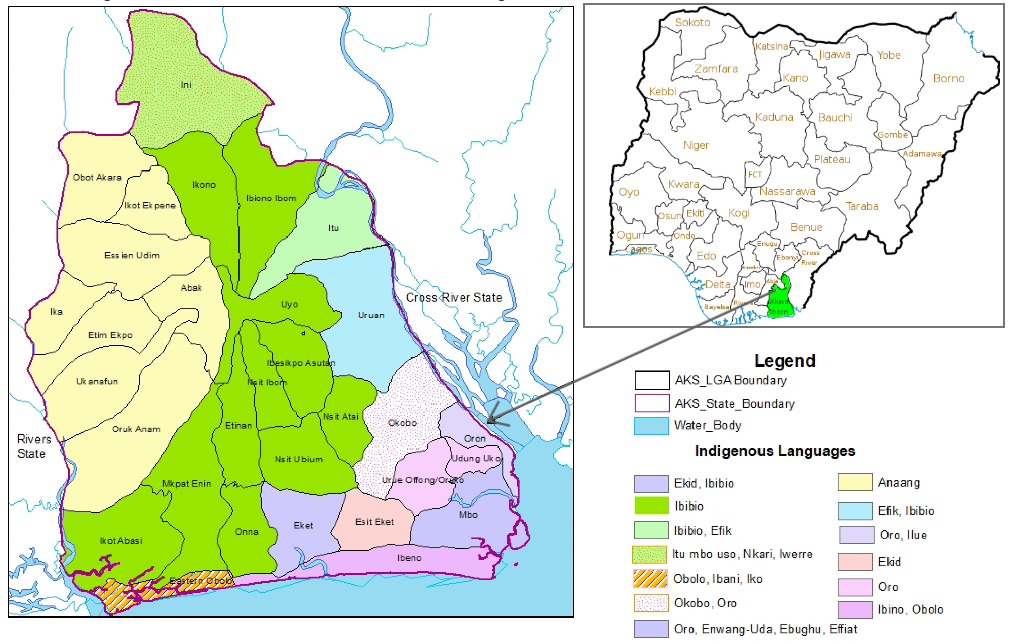}
\centering
		\caption{\textbf{Map showing the regions where the IBOM languages are spoken in Nigeria} including Anaang, Efik, Ibibo, and Oro. The languages are native to the South-South region of Nigeria, primarily spoken in the Akwa-Ibom state painted in Green.}
		\label{AKS}
		\vspace{-1.5em}
\end{figure*}


This work introduces \ibom---a new dataset consisting of two major NLP tasks (machine translation and topic classification) for four ``\textit{minority}'' languages (Anaang, Efik, Ibibio, and Oro) in Nigeria by extending the massively multilingual Flores-200~\cite{nllb2022} and SIB-200~\cite{adelani-etal-2024-sib} datasets. Our evaluation shows that current LLMs perform poorly on machine translation for these languages in both zero-and-few shot settings. We find that leveraging cross-lingual transfer from M2M-100~\cite{fan2021beyond} on religious parallel texts and a related language (Efik) improves performance for the other languages. On topic classification, we find that few-shot prompting steadily improve topic classification performance, and even exceed performance of supervised fine-tuning baseline with AfroXLMR~\cite{alabi-etal-2022-adapting}. Specifically, this work makes the following contributions: 


\begin{enumerate}[label=(\Alph*)]
\item We develop \ibom-MT, a collection of translated sentences from English Flores-200 dataset to four low resource Nigerian languages (i.e. Ibibio, Efik, Anaang, and Oro) that are not represented on Google translate. To our knowledge, this is the first parallel language resource created for Anaang and Oro languages, respectively. 
\item We develop \ibom-TC, a topic classification dataset created by aligning \ibom-MT and SIB-200 topic labels. 
\item We apply several fine-tuned baselines and LLM prompting for machine translation and topic classification on these introduced benchmarks.  
\item We release the datasets from this work and make it available to the research community.~\footnote{\url{https://huggingface.co/collections/howard-nlp/ibom-nlp}}

\end{enumerate}

\section{Related Works}
\paragraph{African machine translation data}
Flores evaluation datasets~\cite{goyal-etal-2022-flores,nllb2022} have emerged as critical resources 
in several low-resource languages.
Beyond the large scale evaluation datasets, there have been several community-driven data collection for Nigerian languages including: MENYO-20k for \yoruba~\cite{adelani-etal-2021-effect}, Igbo-English MT~\cite{ezeani2020igbo},  MAFAND-MT~\cite{adelani2022few} for Hausa and Nigerian-Pidgin, NollySenti~\cite{shode-etal-2023-nollysenti}---a translated benchmark for movie sentiment, and IrokoBench~\cite{adelani-etal-2025-irokobench}---a translated benchmark for knowledge QA, natural language inference and math reasoning. The last two benchmarks only cover Hausa, Igbo, \yoruba and Nigerian-Pidgin. Initiatives like the WMT Shared Task 
have also played an important role in boosting the evaluation of MT on African languages~\cite{adelani2022findings}.

\paragraph{Available Ibom languages data} In the under-studied languages, only Efik has some available bilingual and monolingual data. JW300~\citep{agic-vulic-2019-jw300}---a multilingual parallel corpus based on Jehovah Witness publications contains over 331K parallel sentences with English. Other parallel resources includes SMOL~\citep{caswell2025smol}, and Gatitos dictionary~\citep{jones2023bilex}. Aside bitexts, some monolingual data exists in large filtered Common Crawl data such as GlotCC~\citep{kargaran2024glotcc} and FineWeb-2~\citep{penedo2025fineweb2} but they often smaller in size.

\section{Ibom languages characteristics}

\subsection{Geographical information}
The Ibibio, Efik, Anaang, and Oro languages are predominantly spoken in Akwa Ibom State in the South-South geopolitical region of Nigeria. \autoref{AKS} shows the map of the regions in Akwa Ibom State, Nigeria, where these four languages are spoken. This map indicates the indigenous languages of each Local Government Area in Akwa Ibom State. Ibibio, Anaang, and Oro are shown to be the major languages in Akwa Ibom State as observed by the number of Local Government Areas where the languages are indicated as indigenous languages. In addition, Efik is mainly spoken in the Local Government Areas of Akwa Ibom State which border the Cross River State.
The following is information about each of these languages.

\paragraph{Ibibio} The Ibibio language is primarily spoken in Akwa Ibom State and in some Local Government Areas in Cross River State, in Nigeria. Ibibio serves as the lingua franca of Akwa Ibom State and is spoken as a first language in fifteen Local Government Areas in Akwa Ibom State namely; Uyo, Itu, Uruan, Etinan, Nsit Ibom, Nsit Atai, Nsit Ubium, Ibesikpo Asutan, Ikono, Ini, Ikot Abasi, Mkpat Enin, Ibiono Ibom, Onna, and Eket \cite{Urua2021}. There are approximately 3.7 million speakers of Ibibio \cite{mensah2024representation}.



\paragraph{Efik} The Efik language is spoken in the following local government areas in Akwa Ibom State: Itu, Uruan, and Oron local government area, and Southwest Cameroun. Efik is also spoken in the following local government areas in Cross River State, Nigeria: Calabar Municipality, Calabar South, Akpabuyo, Bakassi, Odukpani, and parts of Akamkpa \cite{offiong2013efik}. Efik has approximately 3.5 million speakers.~\footnote{\url{https://www.britannica.com/}}


\paragraph{Anaang} The Anaang language is primarily spoken in the North West part of Akwa Ibom State, Nigeria. Anaang is predominantly spoken in the following eight Local Government Areas in Akwa Ibom State: Abak, Ikot Ekpene, Essien Udim, Ukanafun, Etim Ekpo, Ika, Obot Akara, and Oruk Anam \cite{udoh2012anaang}. There are approximately 1,400,00 Anaang speakers.~\footnote{\url{https://people.umass.edu/nconstan/representatives/}}


\paragraph{Oro}
The Oro language is predominantly spoken in the following Local Government Areas in Akwa Ibom State: Oron, Urue Offong/Oruko, Okobo, Udung Uko, and Mbo. There are more than 400,000  speakers of Oro.~\footnote{\url{https://www.101lasttribes.com/tribes/oron.html}}


\subsection{Linguistic features}
All the languages belong to the Lower Cross branch of the Cross River Division of the (New) Benue-Congo family~\cite{williamson1989niger,Urua2021}. 
Ibibio, Efik, Anaang belong to the Efik-Ibibo sub-family while Oro is in a different sub-family (Ns\'{i}\textipa{\ng} Oro) 
Phonologically, all the languages are tonal and have a similar tonal system of high, low, downstepped and contour tones \citep{udoh2012anaang, ukpe2018aspects}. The vowels, consonants, and syllable structure of these four languages are presented in \autoref{tab:Phonological}.

Morphologically, Ibibio, Efik, and Anaang are considered agglutinating languages to a large extent and inflecting languages to some extent~\citep{Essien2008, mensah2010prefix, offiong2013efik}. Syntactically, all languages have flexible Subject Verb Object (SVO) sentence structure as shown in the following examples:

\textbf{English}: I am going to school.

\textbf{Ibibio}: ami \~n-ka uf\d{o}k\~nwed.

\textbf{Efik}: ami n-ka uf\d{o}k\~nwed.

\textbf{Anaang}: ami n-ka uf\d{o}k\~ngwed.

\textbf{Oro}: ami n-ga uv\d{o}k\~nwid.

	\begin{table*}[ht]
\centering
\resizebox{\textwidth}{!}{%
\begin{tabular}{>{\centering\arraybackslash}m{1.5cm}|>{\centering\arraybackslash}m{3.5cm}|> {\centering\arraybackslash}m{3.7cm}|>
{\centering\arraybackslash}m{3.5cm}|>
{\centering\arraybackslash}m{4.5cm}}
\hline
\textbf{Language} & \textbf{Vowels} & \textbf{Consonants}& \textbf{Syllable Structure} & \textbf{Tone Pattern} \\
\hline
Ibibio&  a, e, i,  $\underaccent{\cdot}{i}$, o, $\underaccent{\cdot}{o}$, u, $\underaccent{\cdot}{u}$, $\Lambda$, \textschwa & b, d, f, gh, h, k, kp, m, n, $\tilde{n}$, $\tilde{n}$w, ny, p, s, t, w, y & N, V, CV, CGV (CrV), CVC, CVV, CVVC&High(H), Low(L), Downstepped(D), Rising(R), Falling(F) \\
\hline
 Efik &a, e, i,  $\underaccent{\cdot}{i}$, o, $\underaccent{\cdot}{o}$, u,  $\Lambda$ & p, b, d, f, g, h, k, kp, kw, n, ny, ñ, ñy,m, n, p, r, s, t, w, y & N, V, CV, CVC, CGV (CrV), CVV, CVVC&High(H), Low(L), Downstepped (D), Rising (R), Falling (F)\\
\hline
 Anaang & a, e, i, o, $\underaccent{\cdot}{o}$, u, $\underaccent{\cdot}{u}$ & b, ch, d, f, gh, gw, j, k, kp, kw, l, m, n, ñ, ñw, ny, p, r, s, t, w, y &N, V, CV, CVV, CVC, CVVC & High (H), Low (L), Downstepped (D), Rising (R), Falling (F)\\
\hline
  Oro& a, e, $\underaccent{\cdot}{e}$, i,  $\underaccent{\cdot}{i}$, o, $\underaccent{\cdot}{o}$, u& b, d, f, g, gb, gh, gw, j, k, kp, kw, l, m, n, ny, ñ, ñw, r, s, t, v, w, y, z  & N, V, CV, CGV (CrV), CVC, CVV, CVVC &High (H), Low (L), Downstepped (D), Rising (R), Falling (F)\\

\hline

\end{tabular}
}
\caption{\textbf{Phonological characteristics of the Ibom languages}}

\label{tab:Phonological}
\end{table*}

\section{\ibom dataset}

In this work, we used the data from the Flores-200 dataset~\citep{nllb2022}, which is an evaluation benchmark that consists of sentences obtained from English Wikipedia covering various topics and domains. 
We used the data splits from the Flores-200 dataset i.e. DEV (997 sentences) and DEVTEST (1,012 sentences); we also collected 1,000 out of the 6,000 sentences in the NLLB-SEED training dataset. Finally, we extended the DEV and DEVTEST set for topic clasification based on SIB-200 recipe. We provide the data statistics for both tasks in \autoref{tab:data_stat}. 

\begin{table}[t]
 \footnotesize
 \begin{center}
  \begin{tabular}{l|rr}
    \toprule
     \textbf{Split} & \textbf{IBOM-MT (\# sents)} & \textbf{IBOM-TC (\# sents)}  \\
\midrule
Train & 1000 & 701 \\
Dev & 997 & 99 \\
Test & 1012 & 204 \\
\midrule
Total & 3,009 & 1,004 \\
\bottomrule
  \end{tabular}
 \vspace{-2mm}
  \caption{\textbf{\ibom dataset}. We report the data statistics for both  \ibom-MT and \ibom-TC tasks}
  \vspace{-4mm}
  \label{tab:data_stat}
  \end{center}
\end{table}

The lead translator for each of the languages reviewed the translations in their respective languages for errors and made corrections where necessary.

\subsection{\ibom-MT: Machine translation}

For each language of focus in this work (i.e. Ibibio, Efik, Anaang, and Oro), we identified and worked with three (3) linguists who speak, studied, and reside in a State in Nigeria in which these languages are spoken. These linguists were the translators for the dataset collected in this work. Each of these linguists had at least a Bachelors degree in Linguistics. Some of the linguists had a PhD in Linguistics---one is a Full Professor of Linguistics and another is a Lecturer of Linguistics. For each language, one of the translators was the lead translator. Translators received appropriate remuneration.

Given that for each language, we have three translators and there are three sets of data i.e. DEV (997), DEVTEST (1,012), and training (1,000), for each language, one translator translated one set of data. When they were done with the translations, the lead translator for each of the languages reviewed the translations in their respective languages for errors and made corrections where necessary. The translation for this work was done over a period of four months. During this time period, the research team consisting of linguists, NLP experts, and students met weekly to discuss the progress of the project and resolve any issues that arose during the data collection process and afterwards.

\subsection{\ibom-TC (Topic classification) }
We extended SIB-200 topic classification dataset~\citep{adelani-etal-2024-sib} to the Ibom languages by aligning translated sentences with topic labels in English SIB-200. Since, SIB-200 only used DEV and DEVTEST portion of Flores-200 to develop a topic classification, the alignment was straightforward. We used the same script released by the SIB-200 authors to automatically align the translated sentences with English topic labels. Thus, we have \ibom-TC for the four Ibom languages.

\section{Experimental setup}
Here, we describe the experiments used to evaluate the four Ibom-NLP languages. We conduct an extensive evaluation of fine-tuning baselines and LLM prompting on two tasks: machine translation (MT) and topic classification (TC).


\subsection{Fine-tuning baselines}
\subsubsection{Machine translation} We fine-tuned two massively multilingual MT models: M2M-100 (418M parameters)~\citep{fan2021beyond} and NLLB-200 (600M parameters)~\citep{nllb2022}, covering many-to-many translation to/from 100 and over 200 languages, respectively. For ease of fine-tuning, we trained only the smaller parameter versions of the M2M-100 and NLLB-200 models. We fine-tuned the models for 3 epochs using an NVIDIA GH200 120GB GPU machine. We measure the performance of the MT models using three metrics: ChrF++~\cite{popovic-2017-chrf}, BLEU~\cite{papineni-etal-2002-bleu}, and SSA-COMET~\cite{li2025ssa}---an extension of COMET~\cite{rei-etal-2020-comet} embedding-based metric to African languages. 

\paragraph{2-stage MT fine-tuning} 
We explore the 2-stage fine-tuning specifically for the MT task. Given the availability of medium-sized religious parallel data ($\thicksim 331K$ sentences) for English–Efik from MT560~\cite{gowda-etal-2021-many}, we designed a 2-stage fine-tuning approach. In the first stage, we fine-tuned separately on English-to-Efik (\texttt{en-efi}) and Efik-to-English (\texttt{en-efi}) data. In the second stage, we fine-tuned on 1,000 parallel sentences for each language pair.
Since the fine-tuning data is limited, we leverage the effectiveness of cross-lingual transfer to improve performance on related Ibom languages that are not covered in existing multilingual models.

\subsubsection{Topic classification} We fine-tuned six multilingual encoder models: XLM-R~\citep{conneau-etal-2020-unsupervised}, Glot500~\cite{imanigooghari-etal-2023-glot500}, AfriBERTa~\cite{ogueji-etal-2021-small}, Serengeti~\cite{adebara-etal-2023-serengeti}, AfroXLMR~\cite{alabi-etal-2022-adapting}, and AfroXLMR-61L~\cite{adelani-etal-2024-sib}. XLM-R and Glot500 cover 100 and 511 languages, respectively, while the remaining four encoders are African-centric. The multilingual models were created using two approaches: (1) pre-training from scratch, and (2) adapting from pre-trained encoders such as XLM-R via continued pre-training. XLM-R, Serengeti, and AfriBERTa were pre-trained from scratch, although Serengeti covers significantly more languages—primarily from Africa—i.e., 500 vs. 100 (XLM-R) and 11 (AfriBERTa). AfroXLMR was created through multilingual adaptive fine-tuning (i.e., continued pre-training) for 20 widely spoken African languages, while AfroXLMR-61L extends this to 61 African languages. Glot500 first performed vocabulary extension before continued pre-training on 511 languages and has one of the widest coverages of low-resource languages.

\paragraph{\ibom languages coverage in LLMs}
Despite the broad language coverage of some multilingual encoders such as Glot500, AfroXLMR-61L, and Serengeti, only one or two of the Akwa Ibom languages were included during pre-training. Specifically, Glot500 and AfroXLMR-61L include only Efik, while Serengeti includes both Efik and Ibibio. 
Given the linguistic closeness of the Akwa Ibom languages, we believe they can benefit from cross-lingual transfer for both tasks.

\subsection{LLM prompting}
We performed zero-shot and few-shot evaluations on both machine translation and topic classification tasks. We primarily focused on proprietary models, as they have been shown to achieve better overall performance than open models and to provide broader coverage of low-resource languages---for example, Gemini claims to support 400 undisclosed languages~\citep{comanici2025gemini}. We evaluated the GPT-4.1, o4-mini, Gemini 2.0 Flash, and Gemini 2.5 Flash (Thinking) models. For these four models, we conducted zero-shot and few-shot evaluations (5-shot, 10-shot, and 20-shot), where each 5-shot set consists of the first five examples from the training data split.

\section{Results}

\begin{table*}[t]
 \footnotesize
 \begin{center}

  \begin{tabular}{l|ccccc|ccccc||c}
    \toprule
     & \multicolumn{5}{c}{\textit{eng {$\rightarrow$} X}} & \multicolumn{5}{c}{\textit{X {$\rightarrow$} eng}} & \textbf{AVG} \\
    \textbf{Model} &  \textbf{anw} & \textbf{efi} & \textbf{ibb} & \textbf{oro} & \textbf{Avg.} & \textbf{anw} & \textbf{efi} & \textbf{ibb} & \textbf{oro} & \textbf{Avg.} &   \\
\midrule
\multicolumn{12}{l}{\textbf{Encoder-Decoder}}\\
M2M-100FT & 14.7&  12.7&  10.5&  10.7&  12.2&  35.1&  25.0&  23.9&  23.5&  26.9& 19.5\\
NLLB-200 FT &  27.4&  16.7&  15.9&  17.4&  19.4&  32.2&  24.0&  22.8&  23.6&  25.7&  22.5\\
\rowcolor{Gray}
M2M-100 FT: 2-stage &  27.6&  36.2&  24.5&  18.2&  26.6&  37.6&  32.0&  27.7&  24.0&  30.3&  28.5\\
NLLB-200 FT: 2-stage &  22.0&  35.5&  20.6&  18.2 &  24.1&  37.6&  34.6&  29.3&  25.8&  31.8&  28.0\\

\midrule
\multicolumn{12}{l}{\textbf{Decoder-only}}\\
\multicolumn{12}{l}{\textbf{LLM eval: 0-shot}}\\
GPT-4.1  & 25.9 & 23.0 &21.1& 16.1& 21.5& 37.1&27.3 & 26.6& 28.3& 29.8 &25.7\\
o4-mini (\texttt{thinking})  & 26.7& 21.1& 19.4& 10.5& 19.4& 36.2& 26.4& 26.8& 29.0& 29.6 & 24.5 \\
\rowcolor{Gray}
Gemini 2.0 Flash  & 25.8 &  31.1 &  24.2 &  15.1 & 24.1 &  \textbf{38.8} & \textbf{38.6} & \textbf{32.1} &  \textbf{29.2} & \textbf{34.7} & 29.3 \\
Gemini 2.5 Flash (\texttt{thinking})  & 17.7 &  31.4 &  24.3 &  19.6 &  23.3 &  31.8 & 36.2  & 30.7 & 25.2 & 31.0 & 27.1  \\
\midrule
\multicolumn{12}{l}{\textbf{LLM eval: 5-shots}}\\
GPT-4.1  &37.6 & 24.3& 22.4&20.3 &26.1 & 37.6&29.2 & 27.3& \textbf{29.2} &30.8 & 28.5 \\
o4-mini (\texttt{thinking})  & 28.1&20.3 & 19.5& 18.7&21.7 & 34.3&26.0 &26.8 & 28.4 &28.9 & 25.3 \\
Gemini 2.0 Flash  &  \textbf{28.0} &  31.0 & 24.0 & 20.4 & 25.9 &  38.5 & 35.8  & 30.5 &  22.2 & 31.6  & 28.8  \\
\rowcolor{Gray}
Gemini 2.5 Flash (\texttt{thinking})  & 26.0 & 44.4  & 25.2  & 22.2 & 29.5 & 31.8 & 36.2 & 30.7 & 25.2 & 31.0 & \textbf{30.2} \\
\midrule

\multicolumn{12}{l}{\textbf{LLM eval: other-shots}}\\
\rowcolor{Gray}
Gemini 2.5 Flash (\texttt{10-shots}) &  25.6&  \textbf{44.5}&  25.5& \textbf{24.9} & 30.1&  26.7&  35.3&  28.7&  25.2&  29.0& 29.6 \\
\rowcolor{Gray}
Gemini 2.5 Flash (\texttt{20-shots}) &  35.9&   42.6&   \textbf{27.1} &  20.9&  \textbf{31.6} &  11.4&  20.1&  20.4&  20.8&  18.2& 24.9 \\

\bottomrule
  \end{tabular}
  \caption{\textbf{MT Performance of fine-tuned models and LLM prompting on Ibom-MT}. We report ChrF++ score. We highlighted the best result in each experimental setup in \colorbox{Gray}{gray}. }
  \vspace{-4mm}
  \label{tab:main_mt_results_chrf}
  \end{center}
\end{table*}

\begin{table*}[h]
 \footnotesize
 \begin{center}
  \begin{tabular}{l|ccccc|ccccc||c}
    \toprule
    & \multicolumn{5}{c}{\textit{eng {$\rightarrow$} X}} & \multicolumn{5}{c}{\textit{X {$\rightarrow$} eng}} & \textbf{AVG} \\
    \textbf{Model}  &  \textbf{anw} & \textbf{efi} & \textbf{ibb} & \textbf{oro} & \textbf{Avg.} & \textbf{anw} & \textbf{efi} & \textbf{ibb} & \textbf{oro} & \textbf{Avg.} &   \\
\midrule
\multicolumn{12}{l}{\textbf{Encoder-Decoder}}\\
M2M-100 & 5.4& 3.1& 0.8& 3.0& 3.1& 12.7& 5.6& 4.1& 3.8&  6.6& 4.8\\
NLLB-200 & 6.0& 2.0& 1.5& 3.2& 3.2& 8.1& 4.0& 2.7& 2.7&  4.4& 3.8\\
\rowcolor{Gray}
M2M-100 FT: 2-stage &  6.5& 8.8& 2.7& 4.0& 5.5& 14.4& 9.2& 5.2& 3.8&  8.2& 6.8\\
NLLB-200 FT: 2-stage &  2.1& 8.5& 1.3& 1.4& 3.3& 13.2& 12.2& 6.3& 3.4&  8.8& 6.1\\
\midrule
\multicolumn{12}{l}{\textbf{Decoder-only}}\\
\multicolumn{12}{l}{\textbf{LLM eval: 0-shot}}\\
GPT-4.1 & 5.4 & 3.2 & 2.0 &2.5 &3.3 &13.1 &5.5 & 3.7 &5.9   &7.1 & 8.7 \\
o4-mini (\texttt{thinking}) & 4.4& 2.8 & 1.9 & 0.5 & 2.4 &10.4 & 4.5& 3.3 & 5.2 & 5.9 &4.1 \\
Gemini 2.0 Flash  & 5.2& 7.2& 4.0& 3.1& 4.9& 13.5& 13.9& 7.7& 6.4&  10.4& 7.6\\
\rowcolor{Gray}
Gemini 2.5 Flash (\texttt{thinking}) &  2.3& 25.3& 7.7& 4.6& 9.9& 8.9& 20.4& 10.4& 7.0&  11.7& 10.8\\
\midrule
\multicolumn{12}{l}{\textbf{LLM eval: 5-shots}}\\
GPT-4.1 &11.6&3.5 &2.3 & 3.2& 5.1 & 12.0& 5.2& 3.5& 5.2& 6.5 &5.8 \\
o4-mini (\texttt{thinking}) &6.6 &1.3 & 2.2& 2.7&3.2 &8.1 & 4.4&3.5 & 5.2& 5.3 &4.3 \\
Gemini 2.0 Flash  & 5.2& 7.2& 4.0& 3.1& 4.9& 13.5& 13.9& 7.7& 6.4&  10.4& 7.6\\
\rowcolor{Gray}
Gemini 2.5 Flash (\texttt{thinking})  &  2.7& 14.7& 5.2& 4.9& 6.9& 10.2& 15.3& 16.6& 9.3&  12.8& 9.9\\
\midrule
\multicolumn{12}{l}{\textbf{LLM eval: other-shots}}\\
\rowcolor{Gray}
Gemini 2.5 Flash (\texttt{10-shots}) & 12.0& 22.8& 4.4& 6.8& 11.5& 4.7& 17.8& 11.4& 13.0&  11.7& 11.6\\
Gemini 2.5 Flash (\texttt{20-shots}) &  11.0& 11.4& 5.6& 5.2& 8.3& 1.4& 3.9& 3.3& 3.3&  3.0& 5.6\\

\bottomrule
  \end{tabular}
  \caption{\textbf{MT Performance of fine-tuned models and LLM prompting on Ibom-MT}. We report BLEU score. We highlighted the best result in each experimental setup in \colorbox{Gray}{gray}. }
  \vspace{-4mm}
  \label{tab:main_mt_results_bleu}
  \end{center}
\end{table*}

\begin{table}[t]
 \footnotesize
 \begin{center}
  \begin{tabular}{l|rrr}
    \toprule
     & \textbf{Gemini 2.5 Flash} & \textbf{M2M-100} & \\
     \textbf{Language} & \textbf{10-shots)} & \textbf{2-stage FT} & \textbf{Ave.}  \\
\midrule
anw & 9.44 & 31.04 & 20.24 \\
efi & 51.11 & 71.27 & 61.19 \\
ibb & 16.8 & 9.81 & 13.11 \\
\bottomrule
  \end{tabular}
 \vspace{-2mm}
  \caption{\textbf{Human Direct Assessment (DA) score} of the best two MT results}
  \vspace{-4mm}
  \label{tab:human_da}
  \end{center}
\end{table}

\begin{table*}[t]
 \footnotesize
 \begin{center}
 \resizebox{\textwidth}{!}{%
  \begin{tabular}{ll|ccccc|ccccc||c}
    \toprule
    \rowcolor{Gray}
    \textbf{Model} & \textbf{Size} & \multicolumn{5}{c}{\textit{eng {$\rightarrow$} X}} & \multicolumn{5}{c}{\textit{X {$\rightarrow$} eng}} & \textbf{AVG} \\
     \rowcolor{Gray}
     &  & \textbf{anw} & \textbf{efi} & \textbf{ibb} & \textbf{oro} & \textbf{Avg.} & \textbf{anw} & \textbf{efi} & \textbf{ibb} & \textbf{oro} & \textbf{Avg.} &   \\
\midrule
\multicolumn{13}{l}{\textbf{Encoder-Decoder}}\\
M2M-100 & 480M& 6.5& 8.0& -1.2& 0.5& 3.45& 38.9& 32.9& 29.9& 26.2& 32.0 & 17.7\\
NLLB-200 & 600M& 39.9& 43.4& 29.5& 36.3& 37.3& 38.8& 30.3& 29.5& 28.8& 31.9& 34.6\\
M2M-100 FT: 2-stage & &  34.5&  49&  35.3&  24.6&  35.9&  42.1&  41.9&  37.3&  28.0&  37.3&  36.6\\
NLLB-200 FT: 2-stage & &  34.1&  46.1&  38.8&  36.4&  38.9&  44.9&  47.9&  38.8&  31.8&  40.9&  39.9\\
\midrule
\multicolumn{13}{l}{\textbf{Decoder-only}}\\
\multicolumn{13}{l}{\textbf{LLM eval: 0-shot}}\\
GPT-4.1 & -- & 25.8& 26.9& 27.7& 22.5& 25.7& 42.5& 40.1& 41.6& 33.2& 39.4&  32.5\\
o4-mini (\texttt{thinking}) & -- & 27.1& 18.8& 18.1& 19.2& 20.8& 28.1& 28.5& 29.5& 28.5& 28.7&  24.7\\
Gemini 2.0 Flash & -- & 21.9& 25.0& 24.0& 28.1& 24.5& 25.5& 28.3& 27.9& 28.1& 27.5&  26.1\\
Gemini 2.5 Flash (\texttt{thinking}) & -- &  36.2&  48.9&  45.9&  33.3&  41.1&  41.6&  50.4&  45.7&  41.7&  44.9&  43.0\\
\midrule
\multicolumn{13}{l}{\textbf{LLM eval: 5-shots}}\\
GPT-4.1 & -- & 22.1& 26.4& 24.0& 26.6& 24.8& 41.8& 30.7& 32.9& 37.2& 35.7&  30.2\\
o4-mini (\texttt{thinking}) & -- & 17.0& 20.6& 16.8& 19.6& 18.5& 24.2& 25.6& 27.8& 25.6& 25.8&  22.2\\
Gemini 2.0 Flash & -- & 22.1& 24.7& 23.5& 24.5& 23.7& 24.0& 27.4& 26.8& 26.7& 26.2&  25.0\\
Gemini 2.5 Flash (\texttt{thinking}) & -- &  34.5&  38.6&  45.3&  37.9&  39.1&  23.3&  29.5&  31.1&  26.1&  27.5&  33.3\\
\midrule
\multicolumn{13}{l}{\textbf{LLM eval: other-shots}}\\
Gemini 2.5 Flash (\texttt{10-shots}) & -- & 34.1&  50.5&  45.2& 38.8& 42.2&  20.2&  25.2&  26.6&  24.0&  24.0& 33.1\\
Gemini 2.5 Flash (\texttt{20-shots}) & -- &  34.0&   51.0&   44.1&  32.5&  40.4& 20.1&  22.7&  21.6&  20.4&  21.2&  30.8\\

\bottomrule
  \end{tabular}
  }
  \caption{\textbf{MT Performance of fine-tuned models and LLM prompting on Ibom-MT}. We report SSA-COMET score. }
  \vspace{-4mm}
  \label{tab:main_mt_results_comet}
  \end{center}
\end{table*}

\subsection{MT results}
\autoref{tab:main_mt_results_chrf} shows the MT results for fine-tuning baselines and LLM prompting. 

\paragraph{Overall zero-shot poor results by LLMs} 
While some LLMs, such as Gemini, claim to support over 400 languages, we observe extremely low ChrF++ scores for the Akwa Ibom languages—particularly in the en$\rightarrow$xx direction. The performance is slightly better in the xx$\rightarrow$en direction, especially for Anaang (\texttt{anw}), which achieved a ChrF++ score of 37.1 with GPT-4.1 and 38.8 with Gemini 2.0 Flash.
Surprisingly, thinking models such as \ofour{} and \geminiTwoFive{}~\footnote{We set the thinking budget to ``-1"} performed worse than their non-thinking counterparts.

\paragraph{2-stage fine-tuning provides a stronger baseline} 
Fine-tuning with only 1,000 parallel sentences yields very low performance, with ChrF++ scores below 25 in both translation directions. Leveraging cross-lingual transfer by first fine-tuning on 331K en$\rightleftarrows$efi examples, followed by 1K examples, results in a significant performance boost---especially for Efik and related languages (Anaang and Ibibio). Oro showed a more modest improvement compared to the other languages due to being linguistically farther away from the others. We find the 2-stage approach to outperform all zero-shot transfer results, except for the xx$\rightarrow$en direction using \geminiTwo.

\paragraph{Few-shot prompting is more effective for 5-shots and 10-shots} 
We find that 5-shots improve performance across all LLMs. The \geminiTwoFive{} (thinking) model achieved the best overall score based on the ChrF++ metric, outperforming the other LLMs. Thus, we assess the performance of \geminiTwoFive{} with more shots. We observe that 10-shots and 20-shots further improve performance in the en$\rightarrow$xx direction. However, in the xx$\rightarrow$en direction, 20-shots lead to a significant drop in performance. In general, we find few-shot to be more useful for Efik than the other Ibom languages achieving up to 44.5 ChrF++ score, we attribute this to the resource level of Efik since it has more monolingual data than the other languages~\cite{gowda-etal-2021-many}.

\paragraph{Inconsistency in MT metrics for the Ibom languages} 
We find that, at times, ChrF++ and BLEU scores do not fully align. We further evaluated using the African-centric COMET metric~\citep{rei-etal-2020-comet}, SSA-COMET~\citep{li2025ssa} and observed that it appears to be more consistent in the en$\rightarrow$xx direction than in the xx$\rightarrow$en direction. We believe it is important to invest in metric development alongside MT development for low-resource languages. SSA-COMET results are reported in \autoref{tab:main_mt_results_comet}. 

Since the metrics do not fully align, we provide human direct assessment evaluation (Human DA) for some of the results in (\S\ref{sec:humanDA})

\subsection{Human evaluation for MT results}
\label{sec:humanDA}
To verify the automatic metrics, we performed human evaluation on 50 test examples and the MT outputs based on the best two systems identified by ChrF++ in \autoref{tab:main_mt_results_chrf} i.e. Gemini 2.5 Flash 10-shots and M2M-100 FT (2-stage). We only evaluated the Ibom languages to English direction for this evaluation. We collected human direct assessment (DA) using the same tool used by the AfriCOMET~\citep{wang-etal-2024-afrimte}. For each of the languages, we recruited three annotators who are bilingual native speakers of the languages. 

After annotation, we only make use of annotators ratings if it has a high spearman correlation with another annotator, and the spearman correlation is more than $0.5$. Out of the four languages, only Oro did not meet this criteria (<0.2), so, we excluded it from the final evaluation in \autoref{tab:human_da}. For the others, they are between 0.5 and 0.65

\autoref{tab:human_da} shows the Human DA results where most annotators prefer the output of M2M-100 2-stage to that of Gemini 2.5 Flash (10-shots) for the Anaang and Efik languages with more than $+20$ points. However, for Ibibio, annotators slighly prefer Gemini 2.5 Flash. This evaluation highlights the weakness of the current evaluation metrics for many low-resource languages. Surprisingly, we find the human evaluation to slightly correlate with the SSA-COMET~\cite{li2025ssa} evaluation metric in  \autoref{tab:main_mt_results_comet} where SSA-COMET judged Gemini 2.5 Flash to be better than M2M-100 2-stage only for Ibibio, while for other languages, it gave very similar scores for the two models.

\subsection{Topic classification results}

\begin{table}[t]
 \footnotesize
 \begin{center}
 \resizebox{\columnwidth}{!}{%
  \begin{tabular}{l|c|cccc|c}
    \toprule
     \textbf{Model} & \textbf{eng} & \textbf{anw} & \textbf{efi} & \textbf{ibb} & \textbf{orx} & \textbf{Avg.}  \\
\midrule
\multicolumn{5}{l}{\textbf{Encoder models}}\\
XLM-R & \textbf{91.8} & 65.0& 57.5 & 54.9& 46.1& 52.8 \\
AfriBERTa & 80.6& 59.1 & 65.1 & 59.4  & 58.9 & 61.5\\
Serengeti & 86.4& 67.4 & 59.1 & 55.9 & 55.4 & 57.1 \\
Glot-500 & 82.6&  51.8 & 38.2 & 37.9 & 35.2 & 37.1\\
AfroXLMR & 90.7& 64.7& 67.0 & 61.5 & 60.7 & 63.0\\
\rowcolor{Gray}
AfroXLMR-61L & 90.4 & 69.6 & 71.3 & 66.6 & \textbf{66.5} & 68.1\\
\midrule
\multicolumn{5}{l}{\textbf{Decoder-only}}\\
\multicolumn{5}{l}{\textbf{LLM eval: 0-shot}}\\
GPT-4.1 & 89.2 & 60.8 & 44.1 & 42.1 & 50.0 & 49.3 \\
o4-mini & 92.7 & 57.8 & 41.7 & 47.6 & 49.5 & 49.2\\
Gemini 2.0 Flash & 87.7 & 60.3  & 67.6 & 61.8 & 61.8 & 62.9 \\
\rowcolor{Gray}
Gemini 2.5 Flash & 87.8 & 70.1 & 76.5 & 74.0 & 51.0 & 67.9 \\
\midrule
\multicolumn{5}{l}{\textbf{LLM eval: 5-shots}}\\
GPT-4.1 & 87.3 & 56.9 & 45.1 & 41.7 & 47.6 & 47.8 \\
o4-mini & 85.3 & 55.9 & 42.2 & 41.2 & 46.1 & 46.4 \\
Gemini 2.0 Flash & 84.8 &  66.2 &  73.5 & 66.7 & 57.8 & 66.1  \\
\rowcolor{Gray}
Gemini 2.5 Flash &  89.2 &  73.0 & 77.0 &  77.0 & 58.8 &  71.4 \\
\midrule
\multicolumn{5}{l}{\textbf{LLM eval: other-shots}}\\
Gemini 2.5 (10-sh) &  89.2 &  72.6 & 78.4 &  79.4 & 57.4 &  71.9 \\
\rowcolor{Gray}
Gemini 2.5 (20-sh) &  88.7 &  \textbf{73.5} & \textbf{79.4} &  \textbf{80.9} & 65.2 &  \textbf{74.8} \\
\bottomrule
  \end{tabular}
 }
  \caption{\textbf{Topic classification performance of fine-tuned models and LLM prompting on Ibom-TC}. We report accuracy metric. We highlighted the best result in each experimental setup in \colorbox{Gray}{gray}. }
  \vspace{-4mm}
  \label{tab:main_topic_results_}
  \end{center}
\end{table}

\autoref{tab:main_topic_results_} shows the result for the TC using six fine-tuned multilingual encoders and four LLMs. 

\paragraph{African-centric encoders excels the other multilingual encoders} We find that African-centric models such as \afroxlmrSixty{} and \afroxlmr{} achieve significantly better performance than the XLM-R model, which does not support many African languages. Although \afroxlmrSixty{} supports only Efik, it can leverage cross-lingual transfer to improve performance on other Akwa-Ibom languages. While Serengeti covers an additional language, Ibibio, it still performs worse than the AfroXLMR (-61L) variants---likely due to its smaller parameter size or the curse of multilinguality, as it covers 500 low-resource languages.

\paragraph{Fine-tune baselines is better than LLMs in zero-shot settings} 
Overall, we find that the best fine-tuned baseline, AfroXLMR-61L, delivered better overall performance than prompting LLMs in zero-shot settings. However, we find \geminiTwoFive{} to be competitive with fine-tuned models that have seen more than 700 training examples. Nevertheless, there remains a large performance gap compared to the English language, which achieves up to 92.7  points with \ofour{}.

\paragraph{Leveraging few-shot is highly effective for  Gemini LLMs} 
For 5-shot settings, \geminiTwo{} and \geminiTwoFive{} improved performance by $+3.2$ and $+3.5$ points, respectively, over their zero-shot prompting. Similarly, increasing the number of shots to 10 and 20 for \geminiTwoFive{} led to further improvements of $+4.0$ and $+6.9$ points, respectively, compared to the zero-shot result. While the Gemini LLMs show performance gains, we find that \gpt{} and \ofour{} did not benefit significantly from the few-shot examples. This suggests that Gemini is likely more multilingual than the OpenAI models, although further investigation is needed.

\section{Conclusion}
In this paper, we develop new datasets for Akwa-Ibom languages, which are truly low-resource Nigerian languages. While many AfricaNLP papers have focused on the big three national languages of Nigeria---Hausa, Igbo, and Yor\`ub\'a, our paper is one of the first to extend to other low-resource languages in Nigeria. We performed evaluation on both machine translation (a text generation task) and topic classification (a natural language understanding task) by extending Flores-200 and SIB-200 to these languages. Our evaluation shows that LLMs are difficult to adapt for these low-resource languages for the machine translation task, however, we find a more positive adaptation with few-shot prompting for topic classification using \geminiTwoFive LLM. 

In the future, we plan to extend the training data size for the Ibom-MT languages to further boost performance, and to extend COMET evaluation support for these languages. We hope our paper will encourage more investment in NLP beyond the top-10 most spoken languages in Africa. 

\section{Limitations}
Our work has some limitations. In this section we address these limitations.

\begin{enumerate}[label=(\Alph*)]
\item This study focused on four low resourced Nigerian languages i.e. Ibibio, Efik, Anaang, and Oro. While this work has contributed to the development of parallel language resources for these languages, the results from the experiments conducted in this work may not generalize to other languages. 
\item There are more than 500 languages spoken in Nigeria; however, our study covers only four of these languages. We hope that this work and the strategies used for collecting data in these four languages will inform linguists who speak and study the various languages in Nigeria and NLP experts to collect translation data in these languages, conduct NLP research, and share the data and findings from their research with the NLP research community. 
\item For the LLM prompting experiments, we evaluated GPT-4.1, o4-mini, Gemini 2.0 Flash, and Gemini 2.5 Glash (Thinking) models. In the future, we will investigate the performance on other LLMs.

 
\end{enumerate}

\section{Acknowledgments}
David Adelani acknowledges the funding of IVADO and the
Canada First Research Excellence Fund through IVADO R3AI Regroupement Grant. Luel acknowledges the support of Carnegie Corporation of New York provided through AIMS-RIC.

\bibliography{custom}



\end{document}